\documentclass[10pt,twocolumn,letterpaper]{article}

\usepackage{cvpr}
\usepackage{times}
\usepackage{epsfig}
\usepackage{graphicx}
\usepackage{amsmath}
\usepackage{amssymb}

\usepackage{adjustbox}
\usepackage{graphicx}

\usepackage[caption=false]{subfig}

\usepackage{multirow}
\usepackage{textcomp}
\usepackage{subfiles}
\usepackage[table,dvipsnames]{xcolor}
\usepackage{makecell}
\usepackage{xcolor}
\usepackage{soul,cite}

\usepackage[pagebackref=true,breaklinks=true,letterpaper=true,colorlinks,bookmarks=false]{hyperref}

\cvprfinalcopy 

\def\cvprPaperID{****} 

\ifcvprfinal\fi
\begin{document}

\title{End-to-End Lane Marker Detection via Row-wise Classification}


\author{Seungwoo Yoo \quad Hee Seok Lee \quad Heesoo Myeong \\ \vspace{3.px} \quad Sungrack Yun \quad Hyoungwoo Park \quad Janghoon Cho \quad Duck Hoon Kim \vspace{3mm} \\
Qualcomm Korea YH\\
{\tt\small \{yoos,heeseokl,hmyeong,sungrack,hwoopark,janghoon,duckhoon\}@qti.qualcomm.com}
}

\maketitle

\def\algorithmname{E2E-LMD}

\begin{abstract}
In autonomous driving, detecting reliable and accurate lane marker positions is a crucial yet chalprolenging task. The conventional approaches for the lane marker detection problem perform a pixel-level dense prediction task followed by sophisticated post-processing that is inevitable since lane markers are typically represented by a collection of line segments without thickness. In this paper, we propose a method performing direct lane marker vertex prediction in an end-to-end manner, \textit{i.e.}, without any post-processing step that is required in the pixel-level dense prediction task. Specifically, we translate the lane marker detection problem into a row-wise classification task, which takes advantage of the innate shape of lane markers but, surprisingly, has not been explored well. In order to compactly extract sufficient information about lane markers which spread from the left to the right in an image, we devise a novel layer, inspired by~\cite{Drivable}, which is utilized to successively compress horizontal components so enables an end-to-end lane marker detection system where the final lane marker positions are simply obtained via $\textit{argmax}$ operations in testing time. Experimental results demonstrate the effectiveness of the proposed method, which is on par or outperforms the state-of-the-art methods on two popular lane marker detection benchmarks, \textit{i.e.}, TuSimple and CULane.
\end{abstract}

\section{Introduction}

With the explosive growth of the researches and developments on the computer vision technologies with sensor fusion, localization and path planning, the advanced driver assistance system (ADAS) or high-level self-driving system (SDS) has been widely adopted in recent vehicles such as Waymo~\cite{rosenband2017inside}, Uber~\cite{uber-self-driving}, Lyft~\cite{lyft-self-driving}, Mobileye~\cite{yoffie2014mobileye}, Google car~\cite{poczter2014google} and Tesla~\cite{dikmen2016autonomous}. Especially, recent researches and projects~\cite{garnett2017real, dikmen2016autonomous, yoffie2014mobileye} on the ADAS and SDS are focused more on cameras than other sensors, \textit{e.g.} LiDAR, due to the cost, design, and also big accuracy improvements in the camera-based perception systems. Although there are a number of components related to the ADAS or SDS, such as lane marker detection, vehicle detection \& tracking, obstacle detection, scene understanding, and semantic segmentation, lane marker detection is one of the key components in camera perception and positioning for several applications, \textit{e.g.}, lane keeping/change assist.

\begin{figure}[t]
\centering
\includegraphics[width=1.02\linewidth]{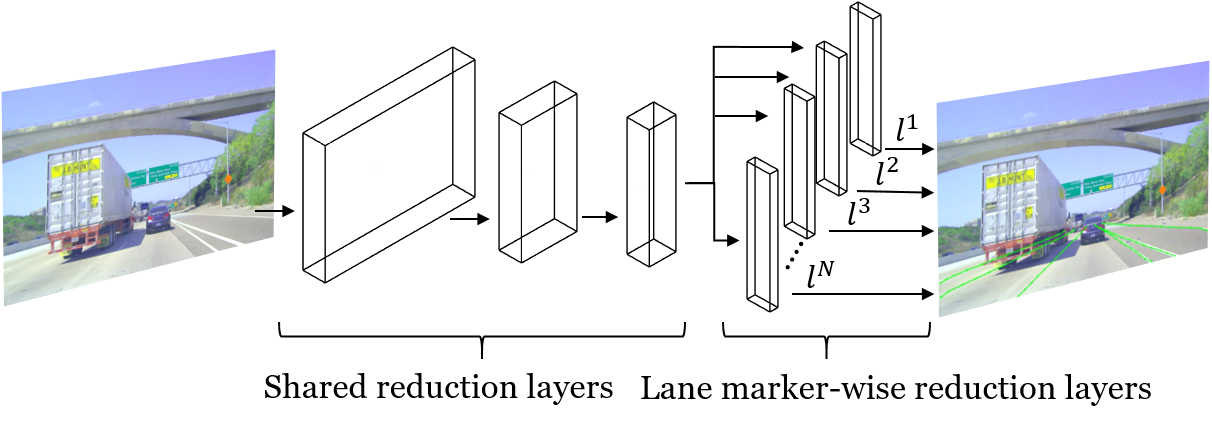}
\caption{The \textbf{\algorithmname} framework for lane marker detection.}
\label{fig:arch_overview}
\end{figure}

A number of researches on lane marker detection have been proposed~\cite{SpatialAsDeep, SAD, LaneGan, Drivable, StixelNet, ElGAN, IV_Instance, ENet, ERFNet, Fitting, MultiRegress, VPG, Clustering}. Most conventional lane marker detection methods are based on two-stage semantic segmentation approaches~\cite{neven2018towards, krahenbuhl2011efficient, hillel2014recent}. In the first stage of these approaches, a network is designed to perform a pixel-level classification to assign each pixel in an image to the binary label, \textit{i.e.}, lane marker or not. However, in each pixel classification, the dependencies or structures between pixels are not specifically considered, and thus additional post-processing is performed in the second stage to explicitly impose the constraints such as uniqueness or straightness of the detected lane marker. The post-processing can be implemented with conditional random field, additional networks, or sophisticated CV techniques like RANSAC, but its computational complexity is not negligible and it should be carefully combined with the first stage by hand-tuning. Therefore, there approaches are hard to scale up for various environments and datasets. Another lane marker detection methods are generative adversarial network (GAN)-based approaches~\cite{LaneGan, ElGAN, luc2016semantic} which considers additional loss to impose such structural constraints.

In this paper, we consider a simple end-to-end framework for recognizing lane marker, called \textit{\algorithmname}, which directly predicts the lane marker vertices without any sophisticated post-processing step (Figure \ref{fig:arch_overview}). Here, the lane marker recognition problem is considered as multiple row-wise classification tasks for each lane marker type where features for classification are expressed through a two-stage module, and the final lane marker positions are simply obtained by $\textit{argmax}$ operations in testing time. The first-stage layers successively compress and model the horizontal components for the shared representation of all lane markers, and the second-stage layers separately model the each lane marker based on this shared representation to directly output the lane marker vertices.

In summary, the contribution of this paper can be summarized as follows: 1) We present a novel and intuitive framework for detecting lane markers. 2) The proposed method is on par or outperforms the recent state-of-the-art methods in both benchmark datasets, \textit{i.e.}, TuSimple and CULane, without complex post-processing. And, finally, 3) We show that the proposed method can effectively capture lane marker representation in an efficient manner with extensive experiments and visualization.  

\section{Related Work}
\label{sec:Related_Work} 

Most traditional lane marker detection methods are based on hand-crafted low-level features. In~\cite{CV_Lane}, they proposed the line segment detection using selective Gaussian spatial filters, which is followed by post-processing steps. Recently, deep learning-based methods are employed to learn to extract features at various scenes. There are mainly two approaches based on convolutional neural networks (CNN): 1) Segmentation-based approach and 2) GAN-based approach.

The first approaches consider lane marker detection as a semantic segmentation task~\cite{SpatialAsDeep, IV_Instance, Fitting, Clustering, SAD}. In~\cite{IV_Instance},  the benefits of lane marker segmentation are combined with a clustering approach designed for instance segmentation. In~\cite{SpatialAsDeep}, they train a spatial CNN (SCNN) with propagating message as residual for detecting long continuous structure. In~\cite{Clustering}, pixel-wise clustering is applied based on conventional segmentation network. In~\cite{Fitting}, authors proposed a deep neural network that predicts a weights map like a segmentation output for each lane marker and a differentiable least-squares fitting module for mapping parameters for curve fitting. In~\cite{SAD}, self-attention distillation (SAD) is proposed to allow the network to exploit attention maps within the network itself and complements the segmentation-based supervised learning.

Second, some methods adopt GAN for lane marker detection tasks. In~\cite{ElGAN}, authors take lane marker labels as extra inputs and use GAN so that the segmentation maps resemble labels to predict the better segmentation outcomes. In~\cite{LaneGan}, they generate low light conditioned images using GAN to increase the environmental adaptability of the network. 

Other deep learning-based methods make an effort to solve lane marker detection from different aspects.  
In~\cite{VPG}, they use extra labels of vanishing point to train the network to output better structural information. In~\cite{MultiRegress}, they consider the lane marker detection and classification problems as regression problems.

One work close to the proposed method is~\cite{Drivable, StixelNet} where column-wise representation is used to recognize free space in road scenes. This horizontal representation for detecting obstacles has been easily utilized for autonomous driving tasks since it can be efficiently translated to an occupancy grid representation. Based on the representation, they used convolutional neural network with simple successive vertical pool layers to regress free space boundaries.  

\def\algorithmname{E2E-LMD}
\def\reductionname{HRM}
\def\reductionnames{HRMs}

\begin{figure*}[t]
	\begin{center}
		\includegraphics[width=1.02\linewidth]{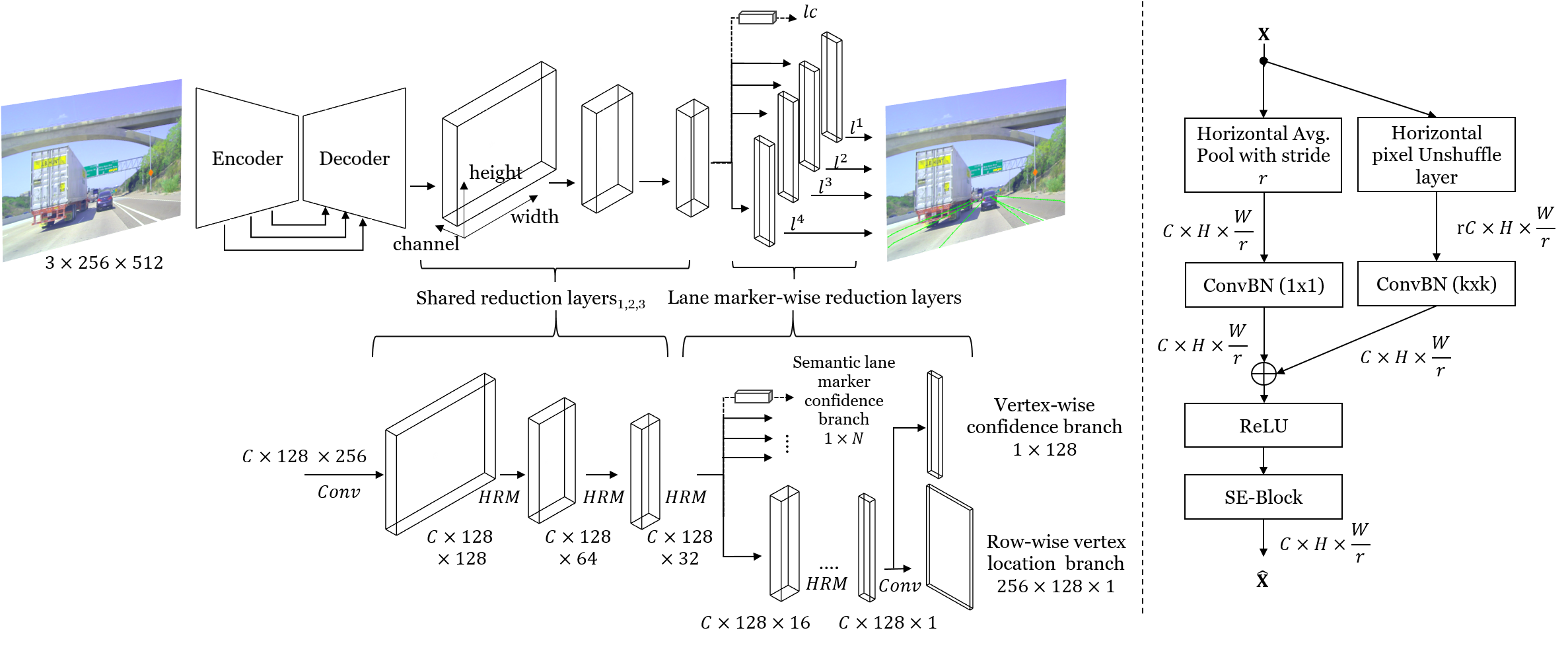}
	\end{center}
	\vspace{-10.pt}
	
	\footnotesize
	\makebox[0.3\textwidth][l]{} (a) The schema of the \textbf{\algorithmname}  \makebox[0.18\textwidth][l]{} (b) The horizontal reduction module (\reductionname)
	
	\vspace{5.pt}
	\caption{The \textbf{\algorithmname} architecture for lane marker detection. We extend general encoder-decoder architectures by adding successive horizontal reduction modules for end-to-end lane marker detection. Numbers under each block denote spatial resolution and channels. \textbf{(a)} Arrows with \reductionname\ denote a horizontal reduction module of (b). Arrows with \textit{Conv} are output convolution with $1\times1$. Dashed arrows denote the global average pooling with a fully connected layer. \textbf{(b)} \reductionname\ is utilized to compress the horizontal representation. $r$ denotes the pooling ratio for width part. Conv kernel size $k$ is set as 3 except the last \reductionname\ layer which set as 1.}
	\vspace{-3.pt}
	\label{fig:arch_deteails}
\end{figure*}

\section{Proposed Method}

As reviewed in Sec.~\ref{sec:Related_Work}, the lane marker detection problem has been tackled with various approaches and each of them has its own pros and cons. However, most of them are based on semantic segmentation with complex post-processing which hinders end-to-end training for extracting lane marker positions.
Inspired by recent works~\cite{Drivable, StixelNet}, we consider the above problem as finding the set of horizontal locations of each lane marker in an image.
Specifically, we divide an image into rows and obtain a row-wise representation for each lane marker using a convolutional neural network. Then lane marker detection can be thought as row-wise classification. In other words, contrasted to the conventional segmentation-based lane marker detection, the proposed method can directly provide lane marker positions.
More specifically, given an input image $\textbf{X} \in R^{3 \times h \times w}$ where $h$ and $w$ are the image height and width, respectively, the objective is to find a lane marker $l_i$ $(i=1, \cdots, N)$ represented by the set of vertices $\{vl_{ij}\} = \{(x_{ij}, y_{ij})\}$ $(j=1, \cdots K)$.
Here, $N$ is the number of lane markers in $\textbf{X}$ which is generally pre-defined, and $K$ is the total number of vertices that is limited to $h$ due to the row-wise representation.


The details of the proposed architecture, which is conceptually simple and can be utilized to any segmentation-based approaches, and its training and inference will be described in the following subsections.

\subsection{Network Architecture}\label{sec:net_arc}

\textbf{Architecture Design:} We propose a novel architecture composed of successive shared and lane marker-wise horizontal reduction modules (\reductionnames), which leads to removing horizontal components spatially and setting the channel size as the target width resolution.

The proposed end-to-end lane marker detection (\textbf{\algorithmname}) architecture consists of three stages (see Fig.~\ref{fig:arch_deteails}(a)). The first stage is a general encoder-decoder segmentation network~\cite{UNet} which encodes information of lane markers in an image and reconstructs spatial resolution. In contrast to standard semantic segmentation approaches, in our implementation, we only recover spatial resolution as the half of an input size to reduce computational complexity. 

In the second stage, we successively squeeze the horizontal dimension of the shared representation using \reductionname s\ without changing the vertical dimension. With this squeeze operation, we can obtain the row-wise representation in a more natural way. After running shared \reductionnames, we squeeze the remaining width of representation by lane marker-wise \reductionnames\ to make single vector representation for each row. We found that it is required to assign dedicated \reductionnames\ on each lane marker after the shared \reductionnames\ for increasing accuracy numbers, since each lane marker has different innate spatial and shape characteristics. For computational efficiency, however, only the first few \reductionnames\ are shared across lane markers, followed by lane marker-wise \reductionnames. With more shared layers we can save computational cost but each lane marker accuracy might be degraded. 

In the last third stage, we have two branches for a lane marker $l_i$: a row-wise vertex location branch and a vertex-wise confidence branch. These branches perform classification and confidence regression on the last \reductionnames features where spatial resolution only has the vertical dimension while the channel size meets the target horizontal resolution $h'$, \textit{i.e.}, $h'=h/2$. The row-wise vertex location branch predicts the horizontal position $x_{ij}$ of $l_i$ per $y_{ij}$ $(y_{ij} = 0, \cdots, h')$.

The vertex-wise confidence branch predicts the existence confidence $vc_{ij}$ whether $(x_{ij}, y_{ij})$ is valid or not. Following~\cite{SpatialAsDeep}, we also add a semantic lane marker confidence branch which produces lane marker-wise existance confidence $lc_i$ after shared \reductionnames.

\textbf{Horizontal Reduction Module:} To effectively compress the horizontal representation, we utilize residual layers proposed in~\cite{Residual} (see Fig.~\ref{fig:arch_deteails}(b)). Specifically, in the skip connection, we add a horizontal average pooling layer with a $1\times1$ convolution to down-sample horizontal components. Although pooling operations let the deeper layers gather more spatial context (to improve classification) and reduce computational complexity, they still have the drawback of reducing the pixel precision. Therefore, to effectively keep and enhance the horizontal representation, inspired by the pixel shuffle layer of~\cite{Shuffle, shuffle2}, we propose to rearrange the elements of $C\times H \times W$ input tensor to make a tensor of shape $rC \times H \times W/r$ in the residual branch, which is somewhat a reverse operation of the original pixel shuffle block in~\cite{Shuffle} so called the horizontal pixel unshuffle layer. By rearranging the representation, we can efficiently move spatial information to channel. Then we apply a convolution operation to reduce the increased channel $rC$ to $C$ which not only reduces computational complexity but also helps to effectively compress lane marker spatial information from the pixel unshuffle operation.

\begin{figure}[t]
	\small
	\begin{center}
		\begin{tabular}{@{}c@{}c@{}}
			\includegraphics[width=.31\linewidth,height=155px]{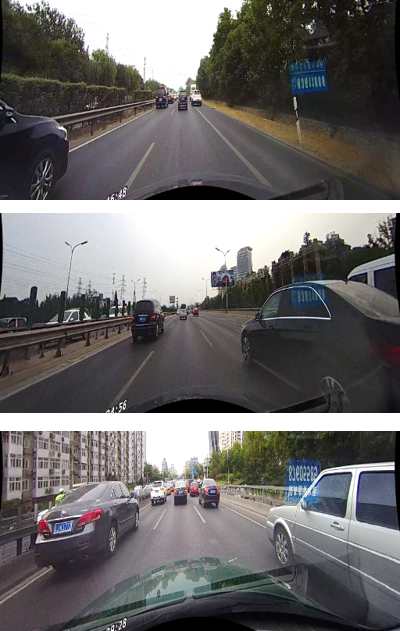} \ & \hspace{.5pt}
			\includegraphics[width=.64\linewidth,height=155px]{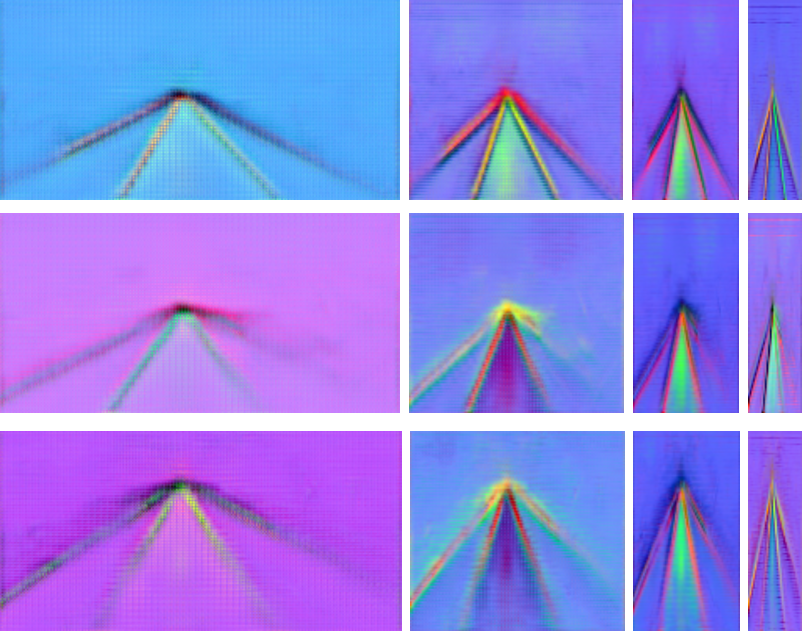} \\
			Input & Learned representations in successive layers
		\end{tabular} 
	\end{center} 
	\caption{\textbf{Learned representations on decoder and shared \reductionname\ layers$_{1,2,3}$:} We visualize how features are encoded in different depths of our shared \reductionname\ layers after decoder. For each layer (row), we visualize the first three principal components as RGB values at each spatial locations. We observe that the features become more distinctive, adapted to specific locations and disentangled in the later layers.}
	\label{fig:lanemarker_vis} 
	\vspace{-5.5px}
\end{figure}

\begin{figure}[t]
	\small
	\begin{center}
		\begin{tabular}{@{}c@{}c@{}c@{}}
			\includegraphics[width=0.5\linewidth,height=40px]{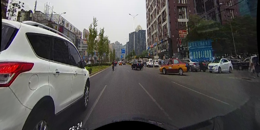} \ & \hspace{.3pt}
			\includegraphics[width=0.22\linewidth,height=40px]{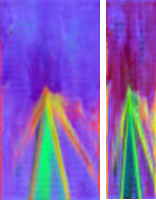} \ & \hspace{.3pt} 
			\includegraphics[width=0.22\linewidth,height=40px]{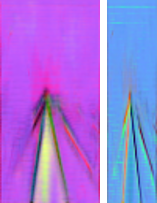} \\
			Input &  Before SE & After SE
		\end{tabular}
	\end{center} 
	\caption{\textbf{Learned representations at shared \reductionname\ layers$_{2,3}$ before/after SE module:} We visualize how encoded features are changed before/after SE block. We observe that after SE block, lane representations become more discernible to be easily separate from each other.}
	\label{fig:lanemarker_vis_se} 
	\vspace{-5.5px}
\end{figure}

To further improve the discrimination between lane markers, we add an attention mechanism by adding Squeeze and Excitation (SE) block~\cite{SEBlock}. The SE block helps to include global information in the decision process by aggregating the information in the entire receptive field and recalibrates channel-wise feature responses which have spatial information encoded by the horizontal pixel unshuffle layer (see Fig.~\ref{fig:lanemarker_vis} and Fig.~\ref{fig:lanemarker_vis_se}).

To confirm the effectiveness of the proposed architecture, we visualize the learned representation using PCA (Principal Component Analysis) (see Fig.~\ref{fig:lanemarker_vis}). The visualized results show that the proposed architecture successfully compress the spatial lane marker information even though we squeeze the horizontal components in representations.

\subsection{Training}
The training objective is to optimize total loss $L$ given by
\begin{equation}
	L = L_{vl} + \lambda_{1} L_{vc} + \lambda_{2} L_{lc},
	\label{eq:loss_all} 
\end{equation}
where $L_{vl}$, $L_{vc}$, and $L_{lc}$ are losses for lane marker vertex location, lane marker vertex confidence, and lane marker-wise confidence, respectively. And $\lambda_{1}$ and  $\lambda_{2}$ are weights for the last two losses. 

\textbf{Lane Marker Vertex Location Loss}:
As we formulated lane marker detection as row-wise classification on lane marker's horizontal position, any loss function for classification can be used.

Specifically, we tested three loss functions, \textit{i.e.}, cross-entropy ($CE$), KL-divergence ($KL$), and PL-loss ($PL$)~\cite{StixelNet}. The $CE$ loss $L_{ij}^{CE}$ for lane marker $l_i$ at a vertical position $y_{ij}$ is computed using the ground truth location $x_{ij}^{gt}$ and the predicted logits $f_{ij}$ having $W/2$ channels. 

To train the lane marker vertex location branch using the $KL$ loss $L_{ij}^{KL}$, we first make a sharply-peaked target distribution of lane marker positions as a Laplace distribution $Laplace_{gt}(\mu, b)$ with $\mu=x_{ij}^{gt}$ and $b=1$, and then compare it with an estimated distribution $Laplace_{pred}(\mu, b)$ by 
\begin{equation}
	\vspace{-2.px}
	\begin{split}
		\mu & =\mathbb{E}_{f_{ij}}[x_{ji}] \\
		& = softargmax(x_{ji})=\sum_{W/2} softmax(f_{ij}) \cdot x_{ij}\\
		b & =\mathbb{E}_{f_{ij}}[|x_{ji}-\mathbb{E}_{f_{ij}}[x_{ji}]|]
	\end{split}
	\label{eq:softargmax}
	\vspace{-2.px}
\end{equation} \\
, similarly with the 2D facial landmark detection algorithm in~\cite{laplace_lm, laplace_gradient_approx}. In case of the $PL$ loss, we follow the original formulation of~\cite{StixelNet} by modeling the probability of lane marker positions as piecewise linear probability distribution.

For an input image, the total lane marker vertex location loss is given by 
\begin{equation}
	L_{vl}=\frac{1}{N} \sum_i^{N} \frac{1}{\sum_j^K e_{ij}} \sum_j^K L_{ij}^{type}\times e_{ij}
\end{equation} \\
, where $type \in \{CE, KL, PL\}$, $e_{ij}$ denotes whether ground truth exists or not, \textit{i.e.}, $e_{ij}=1$ if there is $l_i$ having a lane marker vertex at $y_{ij}$ and $e_{ij}=0$ if not.
\vspace{4.px}

\textbf{Lane Marker Vertex Confidence Loss}:
The lane marker vertex existence is a binary classification problem, thus it can be trained using a binary $CE$ loss $L_{ij}^{BCE}$ between single scalar-value prediction at each $y_{ij}$ location of lane marker $l_i$ and ground truth existence $e_{ij}$. The loss for an entire image is then computed as $L_{ve}=\frac{1}{N \times K} \sum_i^{N} \sum_j^K L_{ij}^{BCE}$.
\vspace{4.px}

\textbf{Lane Marker Label Loss}:
Following~\cite{SpatialAsDeep}, we add a binary $CE$ loss $L_{i}^{BCE}$ to train the lane marker-wise existence prediction. The loss is computed using the predicted $N$-dimensional vector $lc_i$ and existence of each lane $l_i$ in the ground truth. The total loss is then computed as $L_{le}=\frac{1}{N}\sum_i^{N} L_{i}^{BCE}$.
\subsection{Inference}
In testing time, lane marker vertices can be simply estimated per loss as follows: the $\textit{argmax}$ operation is used for the $CE$ or $PL$ loss, and the $\textit{softargmax}$ operation is used for the $KL$ loss.
As mentioned above, there are three outputs from the proposed architecture, \textit{i.e.}, horizontal location of lane marker vertices $x_{ij}$, vertex-wise existence confidence $vc_{ij}$, and lane marker-wise existence confidence $lc_i$. Then the final lane marker $vl_{ij}$ for $l_i$ can be obtained by
\begin{equation}
	\{vl_{ij}\} = \begin{cases}
		\{(x_{ij}, y_{ij}) | vc_{ij} > T_{vc}\} &\text{if } lc_i > T_{lc},\\
		\emptyset &\text{else,}
	\end{cases}
	\label{inference_math}
\end{equation} \\
where $T_{vc}$ and $T_{lc}$ are the thresholds of vertex-wise existence confidence and lane marker-wise existence confidence, respectively. Specifically, the sigmoid output of the vertex-wise and lane marker-wise existence branches is utilized to reject low-confident lane marker vertices and lane markers, respectively.

\def\algorithmname{E2E-LMD}
\def\reductionname{HRM}
\def\reductionnames{HRMs}
\newcommand{\tablestyle}[2]{\setlength{\tabcolsep}{#1}\renewcommand{\arraystretch}{#2}\centering\footnotesize}

\begin{figure}[t]
	\centering
	\includegraphics[width=0.95\linewidth]{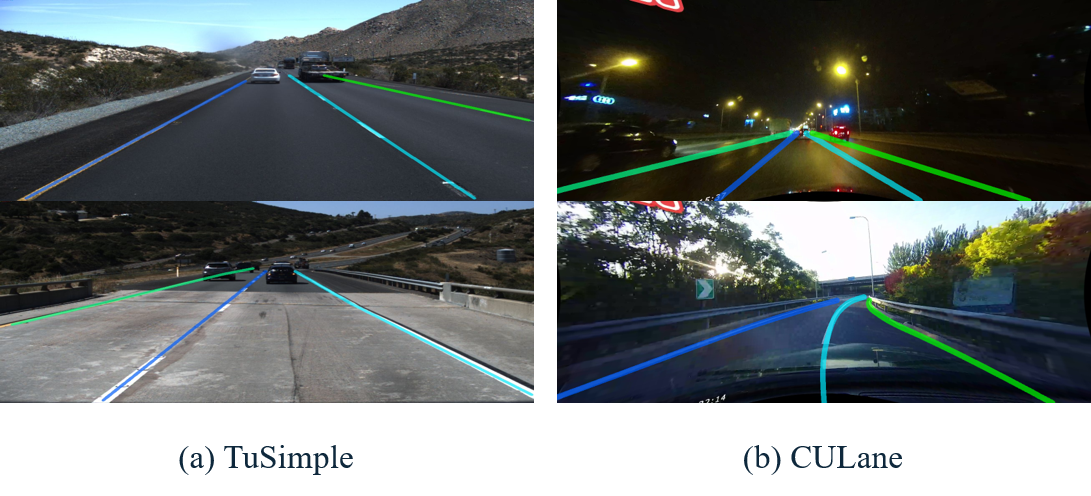}
	\caption{The examples of video frames of (a) TuSimple~\cite{tusimple} and (b) CULane~\cite{SpatialAsDeep}. Ground truth lane markers are shown in various colored lines.}
	\label{fig:dataset}
\end{figure}

\section{Experiments}

\textbf{Datasets:}  We consider two lane marking datasets for evaluating our method. TuSimple~\cite{tusimple} and CULane~\cite{SpatialAsDeep} are widely used in previous works. Some examples of these datasets with ground truth are shown in Fig.~\ref{fig:dataset}.

1) \textit{TuSimple}. The TuSimple dataset consists of 6,408 road images on US highways. The resolution of image is $1280 \times 720$. The dataset is composed of 3,626 for training, 358 for validation, and 2,782 for testing called the TuSimple test set of which the images are under different weather conditions. 

2) \textit{CULane}. The CULane dataset consists of 55 hours of videos which comprise urban, rural and highway scenes, and 133,235 frames are extracted from videos. The dataset is divided into 88,880 frames for training, 9,675 for validation, and 34,680 for testing called the CULane test set. The images have a resolution of $1640 \times 590$. The test set contains 9 different challenging driving scenarios (``Normal", ``Crowd", ``Highlight", ``Shadow", ``Arrow", ``Curve", ``Cross", ``Night" and ``No line").

\textbf{Evaluation Metrics:} For comparing the proposed method with previous lane marker detection methods, we used the following evaluation metrics for each particular dataset:

1) \textit{TuSimple}. We report the official metric used in~\cite{tusimple} as the evaluation criterion. The accuracy is calculated as the average correct number of vertices per image: $Accuracy = \frac{N_{correct}}{N_{gt}}$, where $N_{correct}$ is the number of correctly predicted lane marker vertices, and $N_{gt}$ is the number of ground truth lane marker vertices. Also, we report the false positive ($FP$) and false negative ($FN$) scores.

2) \textit{CULane}. As in~\cite{SpatialAsDeep}, for judging whether the proposed method detects lane markers correctly, we consider each lane marking as a line with 30 pixel width and compute the intersection-over-union (IoU) between ground truths and predictions. Predictions whose IoUs are larger than 0.5 are considered as true positives ($TP$). Then, we used $F_1$-measure as the evaluation metric, which is defined as: $F_{1} = \frac{2 \times Precision \times Recall}{Precision + Recall}$, where $Precision = \frac{TP}{TP+FP}$ and $Recall = \frac{TP}{TP+FN}$.

\vspace{3.px}
\textbf{Implementation Details:} We resized the image of TuSimple and CULane to $256\times512$ and set $N$ as $6$ and $4$ for each dataset. To assign an unique class ID to each lane marker $l_i$, we set labels for each lane marker by ordering the relative distance from an image center. For example, we set the host left lane marker in TuSimple to label $0$, the host right lane marker to label $1$, and the remaining lane markers similarly to cover all $N$ lane markers. For optimization, we used AdamW~\cite{Adamw} with gradual warmup and cosine annealing learning rate schedule with initial learning rate as $8e^{-4}$. The weights $\lambda_1$ and $\lambda_2$ for loss function in Eq.~\ref{eq:loss_all} were set as $10$ and $1$, respectively. The number of shared \reductionname\ was fixed to $3$ for all experiments and the number of channel $C$ was set to $96$. Each mini-batch has 14 images per GPU and we trained using 8 GPUs for 80 epochs on CULane and 140 epochs on Tusimple. Since we only recover the spatial resolution as the half size of an image, we resampled the result vertices to meet the original scale. To reduce overfitting, we applied Dropout with 0.1 probability after every \reductionname. Furthermore, we also applied data augmentation like random cropping, horizontal flipping, and photometric augmentations. In testing time, we set $T_{vc}$, \textit{i.e.}, the threshold of vertex-wise existence confidence, as $0.6$ and $T_{lc}$, \textit{i.e.}, the threshold of lane marker-wise existence, as $0.5$ for every experiment.

\begin{figure}[t]
	\centering
	\includegraphics[width=0.95\linewidth]{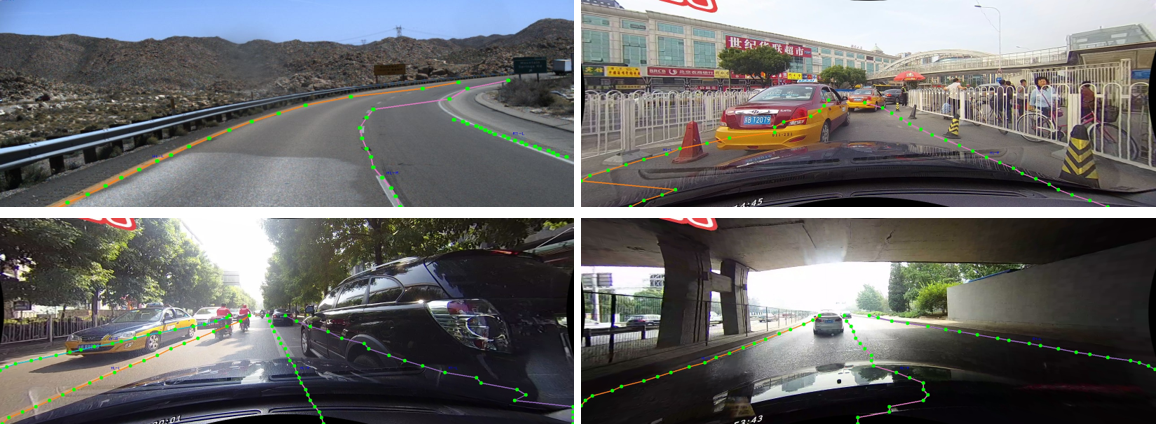}
	\caption{Failed examples from the CULane and TuSimple test sets.}
	\label{fig:results_failed}
\end{figure}
\subsection{Results}

\begin{figure*}[t]
	\centering
	\includegraphics[width=1.\linewidth]{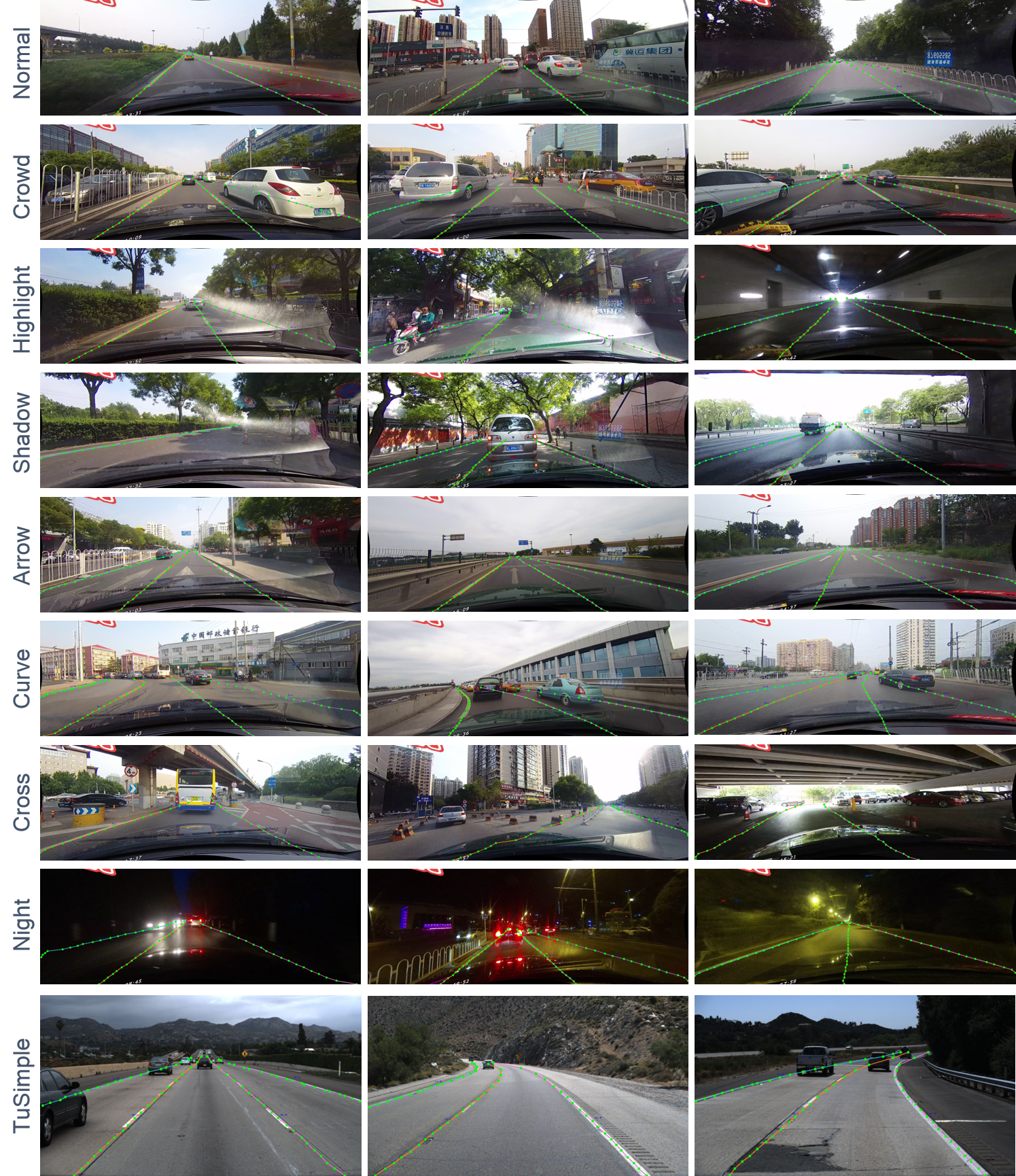}
	The results of \caption{\textbf{\algorithmname} using ERFNet as a backbone network on the CULane and TuSimple test images. All rows except the last one show the CULane test images. Green dots are appropriately sampled for visualization purpose. Best viewed in color.}
	\label{fig:results_all}
	\vspace{1px}
\end{figure*}

\textbf{Quantitative analysis:} To verify the effectiveness of our method, we performed extensive comparisons with several state-of-the-art methods. Following~\cite{SAD}, we evaluated multiple backbones, \textit{i.e.}, ResNet-18 (R-18-E2E), ResNet-34 (R-34-E2E), ResNet-50 (R-50-E2E), ERF (ERF-E2E)~\cite{ERFNet}.  As illustrated in Table~\ref{tusimple_table}, the proposed method attained the competitive performance in the TuSimple dataset. Notable difference compared to other results is low FP ratio, which is obtained without complex post-processing like RANSAC. Interestingly, a heavier network happens to show lower accuracy numbers, \textit{e.g.}, R-34-E2E versus R-50-E2E. The reason would be that the number of the TuSimple training images is not much enough to avoid the overfitting of the network.

In Table~\ref{culane_table}, the proposed method consistently outperforms the state-of-the-art methods in various scenarios of CULane dataset. Especially, the proposed method attained a better performance when comparing~\cite{LaneGan}, which utilizes CycleGAN~\cite{cyclegan} to augment insufficient scenario data.

\textbf{Qualitative analysis:} Fig.~\ref{fig:results_all} shows the localization of lane markers is successful at night, in the shadows, and when passing under the tunnel.
Fig.~\ref{fig:results_failed} shows a few failure cases. The proposed method often fails when there exists reflection over the bonnet that makes it try to find a lane marker and when there are severe curves or occlusions.
\begin{table}[!t]
	\caption{Comparison of different algorithms on the TuSimple test set.}
	\label{tusimple_table}
	\vspace{5px}
	\centering
	\small{
		\begin{tabular}{c|c|c|c}
			\hline
			Algorithm & Accuracy & FP & FN \\
			\hline \hline
			ResNet-18~\cite{SAD} & 92.69\% & 0.0948 & 0.0822 \\
			ResNet-34~\cite{SAD} & 92.84\% & 0.0918 & 0.0796 \\
			LaneNet~\cite{IV_Instance} & 96.38\% & 0.0780 & 0.0244 \\
			EL-GAN~\cite{ElGAN} & 96.39\% & 0.0412 & 0.0336 \\
			FCN-Instance~\cite{Clustering} & 96.5\% & 0.0851 & 0.0269 \\
			SCNN~\cite{SpatialAsDeep} & \textbf{96.53} \% & 0.0617 & \textbf{0.0180} \\
			R-18-SAD~\cite{SAD} & 96.02\% & 0.0786 & 0.0451 \\
			R-34-SAD~\cite{SAD} & 96.24\% & 0.0712 & 0.0344 \\
			\hline \hline
			R-18-E2E &  96.04\% & 0.0311 & 0.0409 \\
			R-34-E2E &  96.22 \% & \textbf{0.0308} & 0.0376 \\
			R-50-E2E &  96.11 \% & 0.0321 & 0.0404 \\
			ERF-E2E &  96.02 \% & 0.0321 & 0.0428 \\
			\hline
		\end{tabular}
	}
\end{table}

\subsection{Ablation Experiments}
We investigated the effects of different choices of our proposed method, \textit{e.g.}, the SE block existence and position, number of shared \reductionname\ layers and loss functions. 

\ \textbf{Architecture:} First, to confirm the pros of including SE block, we evaluated the effect of SE block position and existence on \reductionname\ layer in Table~\ref{tab:exp:ablation_se_position}(a). Following the experiments in the original SE paper~\cite{SEBlock}, we consider three variants: (1) Pre-SE block, in which the SE block is moved before the horizontal pixel unshuffle layer (see Fig.~\ref{fig:arch_deteails}); 
(2) Standard-SE block, in which the SE block is after the residual operation, \textit{i.e.}, after ConvBN in the residual branch; (3) Post-SE block, in which the SE block is moved after the summation of identity connection. Interestingly, in contrast to observations in the original SE paper~\cite{SEBlock}, Post-SE performs much better than other configurations. It seems that the SE block at the end of the residual branch helps to recover the distinctiveness of lane markers whose information could be lost when squeezing the channel in the residual ConvBN layer (see Fig.~\ref{fig:lanemarker_vis_se}). 

\begin{table}[!t]
	\centering
	\caption{Ablation study on different settings}
	\label{tab:exp:ablation}
	\tablestyle{12.pt}{1.1}
	\vspace{5px}
	
	\subfloat[\textbf{SE Position:} Results on the CULane dataset by changing the position of SE block in \reductionname.]{
		\begin{tabular}{c|c|c|c}
			\hline
			ERFNet-E2E & \multicolumn{3}{c}{CULane} \\
			\cline{2-4}
			Architecture & Prec. & Recall & F-measure \\
			\hline \hline
			Without SE & 75.8 & 71.1 & 73.4 \\
			Pre-SE & 75.7 & 71.5 & 73.5 \\
			Standard-SE & 75.0 & 71.6 & 73.3 \\
			Post-SE & \textbf{76.5} & \textbf{71.8} & \textbf{74.0} \\
		\end{tabular}
	}
	\\
	
	\tablestyle{9.pt}{0.9}
	\subfloat[\textbf{Number of sharing pooling layers:} Results on the TuSimple dataset by changing the number of shared \reductionnames.]{
		\begin{tabular}{c|c|c|c|c}
			\hline
			R-18-E2E  & Flops & \multicolumn{3}{c}{TuSimple} \\
			\cline{3-5}
			\# shared & ratio & Accuracy & FP & FN \\
			\hline \hline
			1 & 1.00 & \textbf{96.06}\% & 0.0316 & 0.0436 \\
			2 & 0.56 & 96.05\% & 0.0325 & 0.0419 \\
			3 & 0.34 & 96.04\% & \textbf{0.0311} & \textbf{0.0410} \\
			4 & 0.23 & 95.99\% & 0.0337 & 0.0443 \\
		\end{tabular}
	}\\
	
	\tablestyle{12.pt}{1.1}
	\subfloat[\textbf{Loss function:} Results on the TuSimple dataset by changing the loss function.]{
		\begin{tabular}{c|c|c|c}
			\hline
			R-18-E2E & \multicolumn{3}{c}{TuSimple} \\ 
			\cline{2-4}
			Loss function & Accuracy & FP & FN \\
			\hline \hline
			KL-divergence ($KL$) & 95.49\% & 0.0376 & 0.0551 \\
			PL-Loss ($PL$) & 95.69\% & 0.0455 & 0.0482 \\
			Cross-Entropy ($CE$) & \textbf{96.04}\% & \textbf{0.0311} & \textbf{0.0410} \\
		\end{tabular}
	}\\

	\label{tab:exp:ablation_se_position}
	\label{tab:exp:ablation_pool_sharing}
	\label{tab:exp:ablation_loss_function}
	\label{tab:exp:ablation_base_ch}
\end{table}

\begin{table*}[!t]
	\caption{Comparison of different algorithms on the CULane test set. $F_{1}$-measure is displayed except ``Cross" for which only FP is shown.}
	\label{culane_table}
	\vspace{5px}
	\centering
	\begin{adjustbox}{max width=\textwidth}
		\begin{tabular}{c|c|c|c|c||c|c|c|c|c|c|c|}
			\hline
			Category & \textbf{R-18-E2E} & \textbf{R-34-E2E} & \textbf{R-101-E2E} & \textbf{ERFNet-E2E} & R-18-SAD~\cite{SAD} & R-34-SAD~\cite{SAD} & R-101-SAD~\cite{SAD} & SCNN~\cite{SpatialAsDeep} & ERFNet\cite{LaneGan} \\
			\hline \hline
			
			Normal & 90.0 & 90.4 & 90.1 & 91.0 & 89.8 &	89.9 & 90.7 & 90.6 & \textbf{91.5} \\
			Crowd & 69.7 & 69.9 & 71.2 & \textbf{73.1} & 68.1 & 68.5 & 70 & 69.7 & 71.6 \\
			Highlight & 60.2 & 61.5 & 60.9 & 64.5 & 59.8 & 59.9 & 59.9 & 58.5 & \textbf{66} \\
			Shadow & 62.5 & 68.1 & 68.1 & \textbf{74.1} & 67.5 & 67.7 & 67 & 66.9 & 71.3 \\
			Arrow & 83.2 & 83.7 & 84.3 & 85.8 & 83.9 & 83.8 & 84.4 & 84.1 & \textbf{87.2} \\
			Curve & 70.3 & 69.8 & 70.2 & \textbf{71.9} & 65.5 & 66 & 65.7 & 64.4 & 71.6 \\
			Cross & 2296 & 2077 & 2333 & 2022 & 1995 & \textbf{1960} & 2052 & 1990 & 2199 \\
			Night & 63.3 & 63.2 & 65.2 & \textbf{67.9} & 64.2 & 64.6 & 66.3 & 66.1 & 67.1 \\
			No line & 43.2 & 45.0 & 44.9 & \textbf{46.6} & 42.5 & 42.2 & 43.5 & 43.4 & 45.1 \\
			Total & 70.8 & 71.5 & 71.9 & \textbf{74.0} & 70.5 & 70.7 & 71.8 & 71.6 & 73.1 \\
			\hline

			\hline
		\end{tabular}
	\end{adjustbox}
\end{table*}

\ \textbf{The number of shared \reductionname:} As discussed in Section~\ref{sec:net_arc}, the number of shared \reductionnames\ is an important factor for the speed-accuracy trade-off. We changed the number of shared \reductionnames\ from 0 to 4 (accordingly the number of lane marker-wise \reductionnames\ varies from 6 to 2), and the results are summarized in Table~\ref{tab:exp:ablation_pool_sharing}(b). Note that the batch size of 8 is used in this experiment since the large number of shared \reductionnames\ requires much memory. As shown in Table~\ref{tab:exp:ablation_pool_sharing}(b), we can tune the number of shared \reductionname\ according to the speed-accuracy trade-off.

\ \textbf{Loss function:} To compare loss functions in terms of effectiveness, accuracy numbers per loss function are summarized in Table~\ref{tab:exp:ablation_loss_function}(c). Surprisingly, in our experiments, simple $CE$ loss is preferable than others. The reason would be that the proposed horizontal reduction module helps to effectively incorporate spatial information between the ground truth position and the proximity between neighbors into a network, which leads to helping general $CE$ loss to outperform other specially designed loss functions, \textit{i.e.}, $KL$ and $PL$ losses.

\section{Conclusion} 
In this paper, we proposed a new lane marker detection method to classify each lane marker and obtain its vertex in an end-to-end manner. A novel module for effective horizontal reduction has been devised, and with the module, the state-of-the-art performance is achieved without any complex post-processing. Although we designed the proposed architecture for the lane marker detection problem, it can be also used for other tasks, such as general polygon prediction and semantic/instance segmentation.
 In order to improve the proposed architecture in a better way, we plan to search the reduction module in an automatic manner.

{\small
\bibliographystyle{ieee_fullname}
\bibliography{egbib}
}

\end{document}